\documentclass[10pt,twocolumn,letterpaper]{article}

\usepackage[pagenumbers]{wacv} 

\usepackage{graphicx}
\usepackage{amsmath}
\usepackage{amssymb}
\usepackage{booktabs}
\usepackage{multirow}
\usepackage{float}
\usepackage{svg}
\usepackage{algorithm}
\usepackage{algpseudocode}

\usepackage[pagebackref,breaklinks,colorlinks]{hyperref}

\usepackage[capitalize]{cleveref}
\crefname{section}{Sec.}{Secs.}
\Crefname{section}{Section}{Sections}
\Crefname{table}{Table}{Tables}
\crefname{table}{Tab.}{Tabs.}

\def\wacvPaperID{611} 
\def\confName{WACV}
\def\confYear{2024}

\begin{document}

\title{Army of Thieves: Enhancing Black-Box Model Extraction via Ensemble based sample selection}

\author{Akshit Jindal\\
IIIT-Delhi\\
{\tt\small akshitj@iiitd.ac.in}
\and
Vikram Goyal\\
IIIT-Delhi\\
{\tt\small vikram@iiitd.ac.in}
\and
Saket Anand\\
IIIT-Delhi\\
{\tt\small anands@iiitd.ac.in}
\and
Chetan Arora\\
IIT Delhi\\
{\tt\small chetan@cse.iitd.ac.in}
}
\maketitle

\begin{abstract}
Machine Learning (ML) models become vulnerable to Model Stealing Attacks (MSA) when they are deployed as a service. In such attacks, the deployed model is queried repeatedly to build a labelled dataset. This dataset allows the attacker to train a thief model that mimics the original model. To maximize query efficiency, the attacker has to select the most informative subset of data points from the pool of available data. Existing attack strategies utilize approaches like Active Learning and Semi-Supervised learning to minimize costs. However, in the black-box setting, these approaches may select sub-optimal samples as they train only one thief model. Depending on the thief model's capacity and the data it was pretrained on, the model might even select noisy samples that harm the learning process. In this work, we explore the usage of an ensemble of deep learning models as our thief model. We call our attack Army of Thieves(AOT) as we train multiple models with varying complexities to leverage the crowd's wisdom. Based on the ensemble's collective decision, uncertain samples are selected for querying, while the most confident samples are directly included in the training data. Our approach is the first one to utilize an ensemble of thief models to perform model extraction. We outperform the base approaches of existing state-of-the-art methods by at least 3\% and achieve a 21\% higher adversarial sample transferability than previous work for models trained on the CIFAR-10 dataset. Code is available at: \url{https://github.com/akshitjindal1/AOT_WACV}.
\end{abstract}


\section{Introduction}
\begin{figure}[t]
    \centering
    \includegraphics[width=\linewidth]{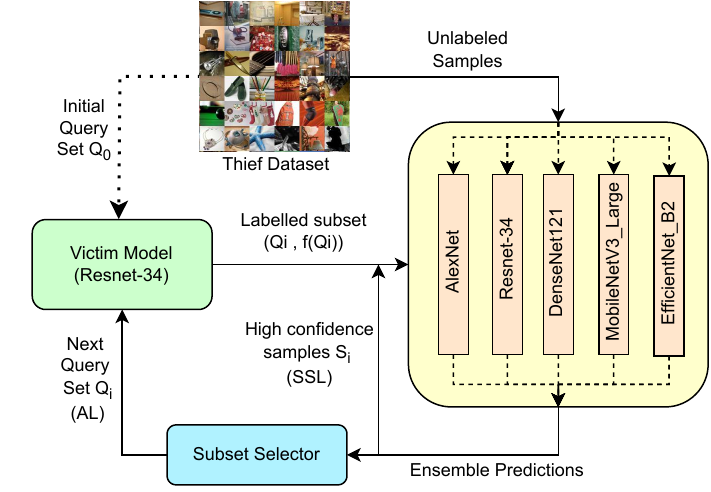}
    \caption{Overview of the Army of Thieves (AOT) Extraction Pipeline. The initial query set $Q_0$ is selected at random. Query outputs are then used to train the ensemble of five thief models. Based on the prediction confidence, highly confident samples are directly added to the labelled subset (Semi-Supervised Learning), while others are sent to the subset selector to curate the query set for the next training cycle (Active Learning).}
    \label{fig:basic_image}
\end{figure}

Machine Learning (ML) models have become essential in various industries because they can learn from data and make accurate predictions. Many companies train and deploy models using cloud-based services such as Google Cloud Platform, Amazon web services etc. These services provide access to ML tools and infrastructure at a reasonable cost, making it viable to utilize ML commercially. A trained model's predictions are made available to the general public via an Application Programming Interface (API), where each API call incurs a cost to the user. This allows the model owner to monetize their trained model without exposing its inner workings. However, recent studies \cite{tramer2016stealing,wang2022black,pal2020activethief,papernot2017practical} show that a malicious user can steal the deployed model's functionality even in such scenarios by querying it on a selected set of inputs. These types of attacks are called Model Extraction or Model Stealing attacks and constitute a significant threat to the intellectual property of the model owner. Moreover, stealing a model's functionality makes other attacks, such as membership inference \ and adversarial attacks, easier.

The model exposed to the risk of being stolen is referred to as the ``victim'' model, while the replica developed by the attacker is known as the ``thief'' model. The attacker aims to create a model with a comparable input-output behaviour to the victim's model. To accomplish this, the attacker assembles a pool of labelled or unlabeled data that is semantically similar or dissimilar to the victim's training data. Queries to the victim model are made using samples from this data pool, and the resulting predictions are utilized for training the thief model. The images in the attack dataset may be publicly available or artificially generated. Due to the limited resources of the attacker, it is not feasible to query every collected image blindly; thus, the attacker must carefully choose the images for each query. Researchers have used methods such as Active Learning \cite{pal2020activethief} and Reinforcement Learning\cite{orekondy2019knockoff} to make each query as informative as possible, allowing the attacker to select the most useful subset of samples to query the victim efficiently. After exhausting the query budget, Semi-supervised learning techniques have been employed to effectively utilize the remaining unlabeled dataset \cite{wang2022black}.

The intelligent sample selection strategies have limitations in the black-box setting, where the attacker cannot access the model's architecture, training hyperparameters, and/or training data. In such a scenario, the attacker can query the model and observe only the output prediction, making it challenging to select informative samples.
The amount of training data required to train a thief model rises with the rise in complexity of the thief model's architecture. This also leads to suboptimal sample selection in the first few cycles of the active learning process, as the model cannot accurately capture the required information from a low number of samples. As a result, previously proposed attacks require a large query budget before they can capture relevant information to train the thief model. 

In this study, we investigate the utilization of an ensemble of thief models for conducting model extraction attacks. An overview of our approach is given in Figure \ref{fig:basic_image}. Ensembling techniques leverage the collective intelligence of multiple models to improve overall performance. By combining the knowledge and predictions of multiple thief models, we can alleviate individual model limitations, reduce noise and uncertainty, and ultimately increase the extraction success rate. This approach holds promise for addressing challenges associated with model extraction, such as intricate model structures, defensive mechanisms, and noise in the extracted information. However, due to the constraint of maintaining query costs within a budget, selecting samples that facilitate improved learning for each member of the ensemble becomes challenging. Furthermore, transferring existing Semi-Supervised Learning (SSL) techniques to the ensemble approach is not a straightforward task.

To enhance diversity, we incorporate ensemble members of different sizes and complexities. Each member is trained on the same set of labelled samples; the ones already queried from the victim model. Sample selection is based on the disagreement observed among ensemble members, as it serves as a metric for quantifying sample complexity. Specifically, samples with high disagreement are considered more valuable for training since they pose greater classification challenges. This approach offers several advantages, including reduced bias towards a single thief model and the selection of samples that better capture the joint input-output distribution. Additionally, due to budget constraints, the model extraction pipeline resembles Few-shot learning methods, thereby increasing the risk of overfitting a single thief model on limited available data. Conversely, the ensemble learning approach acts as a regularization technique, mitigating overfitting. Moreover, the output labels of an ensemble are less noisy and have higher confidence than a single model. The higher confidence leads to better semi-supervised learning once the query budget is depleted. Our work is the \textit{first} to explore the usage of an ensemble of thief models for performing a model extraction attack. We will release the code and pretrained models upon acceptance.

\begin{figure*}[t!]
    \centering
    \includegraphics[width=\linewidth]{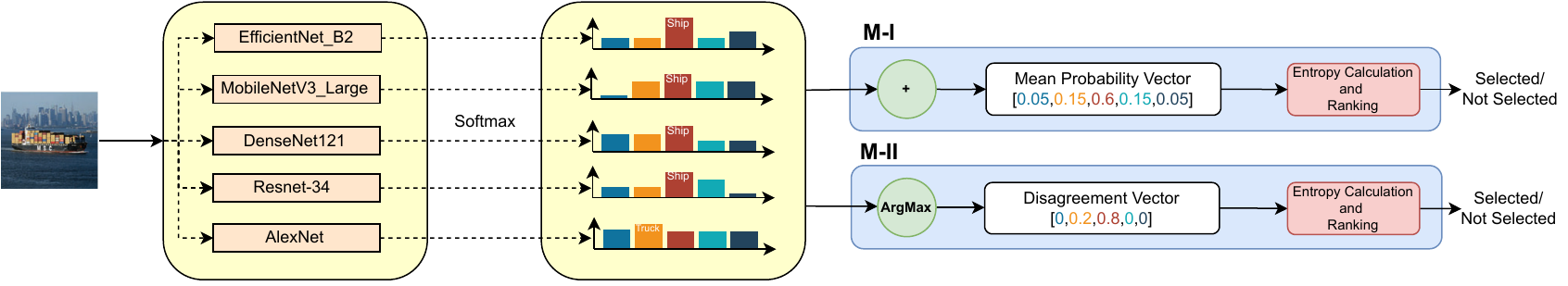}
    \captionsetup{justification=centering}
    \caption{Detailed diagram of the Subset Selector module used for Active Learning. \textbf{M-I}: The mean Probability vector is computed for the Consensus Entropy method. \textbf{M-II}: Entropy of the Disagreement vector is calculated to measure Label Disagreement. Both methods are independent; only one is used for sample selection in a particular experiment.}
    \label{fig:subset_selector}
\end{figure*}



\section{Related Work}
We study existing attacks in the following aspects:
\subsection{Attack Objective} 
While the primary objective of model extraction is to create a duplicate copy of the target or victim ML model, the secondary objective might differ. The three secondary objectives, according to Jagielski et al. \cite{jagielski2020high}, are a) accuracy close to that of the victim model \textbf{(Task accuracy)}, b) prediction agreement over the victim's private training data \textbf{(Fidelity)} and c) prediction agreement over the entire input domain \textbf{(Functional Equivalence)}. We propose that an ensemble-based approach to model stealing would cover all three objectives as each individual model targets accuracy and fidelity while the overall ensemble would lead to functional equivalence.

\subsection{Attack Strategy} 
Tramer et al. \cite{tramer2016stealing} were the first to steal ML models via prediction APIs. Following their work, many model extraction attacks have been proposed over the years. These attacks can be broadly classified into two categories: Exact extraction and Equivalent extraction. Exact extraction \cite{tramer2016stealing, oh2019towards} aims to recover the exact parameters of the victim model, e.g., equation solving to recover a linear regression model. However, extracting exact parameters for neural networks is impossible due to the sheer number of parameters and also due to the fact that different sets of parameters lead to similar outputs. Therefore, learning-based approaches \cite{pal2020activethief, papernot2017practical, sun2022exploring, chandrasekaran2020exploring} have been proposed for equivalent extraction. These approaches treat the victim model as an oracle and treat a separate model using the model's outputs in a supervised learning fashion. Our ensemble-based approach falls under the equivalent extraction category.

\subsection{Query Output from Victim}
The amount of information provided by the victim model directly correlates to the extraction accuracy. Previous works that utilize the complete probability vector \cite{pal2020activethief, orekondy2019knockoff, truong2021data} perform far better than approaches that work in the top-1 or hard-label \cite{correia2018copycat, papernot2017practical, wang2022black, sanyal2022towards} setting. However, the hard-label setting is closest to a real-world scenario as APIs often provide access only to the predicted label. In our work, we utilize only the predicted hard label from the victim. 

\subsection{Query dataset used}
To perform queries to the victim model, previous works have utilized adversarial sample generation \cite{papernot2017practical, hong20200wn}, synthetic sample generation via GANs \cite{zhou2020dast, truong2021data, sanyal2022towards}, and publicly available Out-of-Distribution datasets \cite{pal2020activethief, orekondy2019knockoff, wang2022black, he2021model}. We utilize the publicly available dataset ImageNet\cite{deng2009imagenet} in our work for stealing Image Classification models.

\subsection{Learning Paradigms used}
Learning-based model extraction methods treat the victim model as an oracle. Various attacks utilize Active Learning\cite{pal2020activethief, chandrasekaran2020exploring, jagielski2020high} and Reinforcement Learning \cite{orekondy2019knockoff} to improve query efficiency. Moreover, Semi-supervised approaches have been used for Consistency Regularization \cite{papernot2017practical,wang2022black} after the query budget is exhausted to use the remaining data samples. So far, there have been no approaches that utilize an ensemble of thief models for performing model extraction though there have been approaches that steal an ensemble of models\cite{10007048}.


\section{Proposed Attack}
\begin{figure*}[t!]
    \centering
    \includegraphics[width=\linewidth]{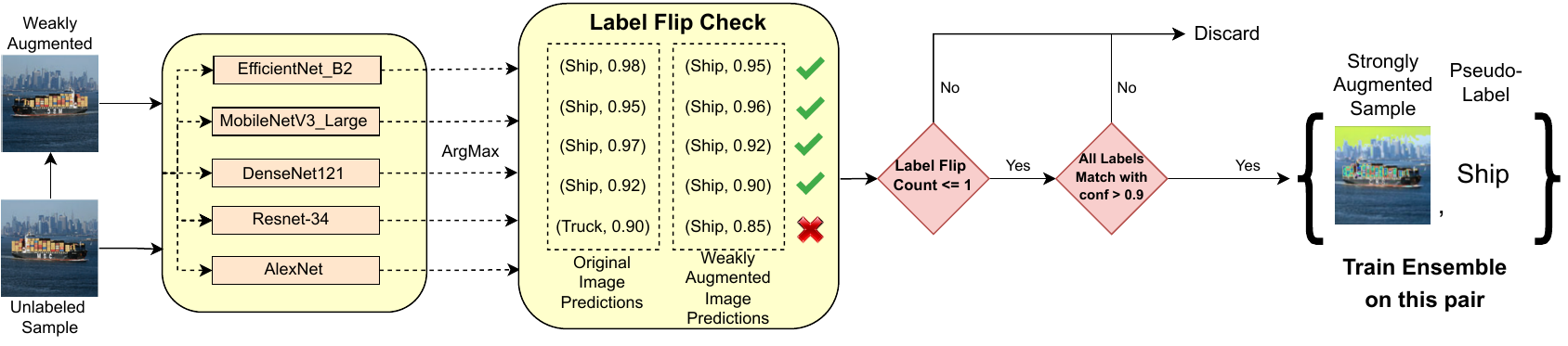}
    \captionsetup{justification=centering}
    \caption{Detailed diagram of the Semi-Supervised Learning pipeline. Weak Augmentations like RandomHorizontalFlip are applied to each unlabeled sample, and the number of label changes are counted. Samples are selected iff the labels of at least 4 models do not change and each label is the same with a minimum confidence value of 0.9. The selected samples undergo a strong augmentation like RandAugment and the ensemble is trained on this (sample,pseudo-label) pair.}
    \label{fig:ssl}
\end{figure*}

\subsection{Recruiting Soldiers} 
It is a well-known fact that using an ensemble of machine learning models yields better results than using a single model, as the ensemble can handle noisy data and prevent overfitting. Thus we pick five models that have been widely accepted as standard architectures for image classification. All of the selected models are pretrained on the ImageNet dataset and are publically available via the Pytorch model hub. These models vary in their amount of trainable parameters and overall learning capacity, measured by their top-1 accuracy on the ImageNet dataset. However, we keep one model architecture the same as the victim model for comparison purposes.

\subsection{Training the army}
\textbf{Active Learning} Following the conventional model extraction pipeline, a random set of samples ($Q_0$) is selected from the unlabeled set ($D_{UL}$) for querying the victim model ($f$) in the first cycle. Each ensemble member ($E_i$) is then trained on the returned labels ($f(Q_0)$) in a supervised learning fashion. After each training cycle, the next set of unlabeled samples ($Q_i$) for querying the victim model is chosen in one of two ways:
\begin{itemize}
    \item \textbf{Consensus Entropy} - For each unlabeled sample $D_{{UL}_i}$ $\in$ $D_{UL}$, the mean probability vector is computed from the softmax output of each ensemble member ($p_{{E}_i}$) to create a consensus vector ($p_{cons}$). Then the entropy of this consensus vector serves as the uncertainty measure for that sample. Samples with the highest consensus entropy are queried in the next cycle.
    \begin{equation}
        H(p_{cons}) =  - \sum_{i=1}^{N} p_{{cons}_i} \log(p_{{cons}_i}) = H (\sum_{i=1}^{5} p_{{E}_i})
    \end{equation}
    Here N is the number of classes in the victim's output space and H is the entropy function.
    
    \item \textbf{Label disagreement} - For each unlabeled sample, the disagreement between ensemble members' outputs serves as the metric for sample selection. If each ensemble member outputs a different label for a sample, it is likely a difficult sample and thus should be selected for querying. For each sample, we gather the output labels and create a label distribution vector $p_{dis}$. The entropy of this vector tells us the level of disagreement, as the entropy will be zero if all members output the same label and will be highest when each output label is different.
    \begin{equation}
        H(p_{dis}) = H ([p_{c_1},p_{c_1},...,p_{c_N}])
    \end{equation}
    where,
    \begin{equation}
        p_{c_i} = \frac{{\text{{Count of Label i in ensemble outputs}}}}{{\text{{Number of ensemble members}}}}
    \end{equation}
\end{itemize}

A detailed diagram explaining both selection strategies with an example is given in Figure \ref{fig:subset_selector}. The samples selected at the end of the active learning cycle become part of the labelled subset $D_{L}$, i.e. \[
D_L = \bigcup_{k=1}^{cycles} Q_k
\] The number of cycles is usually kept equal to 10. Each sample belongs to the ImageNet dataset, but the labels from the victim model replace the corresponding output labels. \\

\textbf{Semi-Supervised Learning} After the query budget has been exhausted, we utilize the remaining unlabeled data via Semi-supervised learning. This allows us to further improve the accuracy of the thief models without querying the victim. Our semi-supervised approach is inspired by FixMatch \cite{sohn2020fixmatch}, as shown in Figure \ref{fig:ssl}. A concise version of our sample selection algorithm is provided in Algorithm \ref{alg:ensemble_semisupervised}. The exact steps are as follows:
\begin{enumerate}
    \item First, we apply a weak augmentation, such as a Random horizontal flip, to our entire unlabeled set $D_{UL}$ (nearly a million images). We infer each original sample $\mathbf{x}$ and its weakly augmented counterpart $\mathbf{x'}$ from our trained AOT and store the output probability vectors $\mathbf{p_i}$ for each ensemble member. 
    \item Secondly, we remove any samples for which the labels of one or more ensemble members change after weak augmentation, i.e. $argmax(p_i(\mathbf{x})) \neq argmax(p_i(\mathbf{x'}))$. The change shows a lack of robustness in the ensemble's predictions for such samples, and we don't want to reinforce such knowledge. 
    \item Then we filter those samples for which all ensemble members unanimously agree upon the output label $\mathbf{y}$. Furthermore, we select only those samples for which each model's output label confidence is above a certain confidence threshold, i.e. max($\mathbf{p_i}$) $\geq$ threshold. 
    \item The chosen samples $\mathbf{x}$ and their corresponding pseudo-labels $\mathbf{y}$ are then utilized for training the ensemble. Like the FixMatch approach, we apply strong augmentations such as RandAugment \cite{cubuk2020randaugment} to each sample before input and train the model to output the pseudo-labels. 
    \item To prevent forgetting, the learning rate is set to a small value of the order 1e-3, and each epoch consists of traversing both the labelled subset and the pseudo-label set once. The losses for each set are combined as per equation \ref{eq:loss} and subsequently backpropagated for training the thief models.
    \begin{equation}
        \text{Total Loss} = Loss (D_L) + \lambda* Loss (D_{pseudo})  \label{eq:loss}
    \end{equation}
\end{enumerate}

To sum up, we use the ensemble's collective decision to initially select samples for querying and then select the most confident samples for consistency regularization. The former extracts the maximum amount of accuracy possible and the latter helps with the generalization ability of each ensemble member.
\begin{algorithm}[H]
\caption{Sample Selection for Semi-supervised Learning}
\label{alg:ensemble_semisupervised}
\begin{algorithmic}[1]
\Require Unlabeled set $D_{UL}$, Ensemble members $E$, confidence threshold $\textbf{t}$
\State Train the ensemble of models $E = \{E_1, E_2, \ldots, E_k\}$ on labeled set $D_L$ with cross-entropy loss.
\For{each sample $\mathbf{x}$ in $D_{UL}$}
    \State $\mathbf{x'} = \text{WeakAugment}(\mathbf{x})$ 
    \For{each ensemble member $E_i$ in $E$}
        \State $p_i(\mathbf{x}) = E_i(\mathbf{x})$
        \State $p_i(\mathbf{x'}) = E_i(\mathbf{x'})$
        \State $l_i(\mathbf{x}) = \text{argmax}(p_i(\mathbf{x}))$
        \State $l_i(\mathbf{x'}) = \text{argmax}(p_i(\mathbf{x'}))$
    \EndFor
    \State $\textbf{changes} = \sum_{i=1}^{k} \mathbb{I}\{l_{i}(\mathbf{x}) \neq l_{i}(\mathbf{x'})\}$ 
    \If{$\textbf{changes} \leq 1$}
        \If{$\forall i, l_i(\mathbf{x}) = y$ and $p_i(l_i) \geq t$}
            \State Add sample ($\mathbf{x}$, $\mathbf{y}$) to $D_{pseudo}$
        \EndIf
    \EndIf
\EndFor
\end{algorithmic}
\end{algorithm}


\section{Experiments}
\subsection{Victim Models}
We train the victim Resnet-34\cite{he2016deep} model on four public image classification datasets: CIFAR-10\cite{krizhevsky2009learning}, CIFAR-100\cite{krizhevsky2009learning}, Caltech-256\cite{griffin2007caltech} and CUBS-200\cite{wah2011cubs} to study and compare the proposed attack's effectiveness. We choose the above architecture and datasets as they have been used for similar studies in the past. The test accuracies for each model are 92.18\%, 61.38\%, 78.43\%, and 77.11\%, respectively. All models were trained using an SGD optimizer with a momentum of 0.5 for 200 epochs. We started with a base learning rate of 0.1, which was decayed by a factor of 0.1 every 30 epochs. After training, all models are treated as black-box APIs, i.e., we only utilize the final one-hot predictions for every input image.

\subsection{Thief Model architectures}
Previous works utilize the same model as the victim or a smaller model as the thief model. In our work, we use an ensemble of five models, namely Resnet-34 \cite{he2016deep}, Alexnet\cite{alexnet}, MobilenetV3\cite{koonce2021mobilenetv3}, Densenet121\cite{densenet} and EfficientNet\_B2\cite{tan2019efficientnet}, as our thief model. The model sizes and the number of parameters are given in Table \ref{tab:arch-comp}. All members are pretrained on the Imagenet dataset. One ensemble member, i.e., Resnet-34, was kept the same as the victim model. Other members were chosen based on their size, learning capacity, and architecture to enforce diversity. 

Alexnet has the lowest learning capacity due to its simpler architecture despite having the largest number of trainable parameters. On the other hand, EfficientNet has the highest learning capacity, as is evident by its performance on the ImageNet dataset. MobileNet has a relatively lower learning capacity but is useful in low-resource environments. DenseNet has a moderate number of parameters and is able to capture complex patterns in the data. Thus, each member contributes to the diversity of the ensemble in different ways. 
\begin{table}[]
\centering
\begin{tabular}{lccc}
\hline
\textbf{Model Arch.} &
  \textbf{\begin{tabular}[c]{@{}c@{}}Size \\ (MB)\end{tabular}} &
  \textbf{\begin{tabular}[c]{@{}c@{}}Parameters \\ (millions)\end{tabular}} &
  \textbf{\begin{tabular}[c]{@{}c@{}}ImageNet \\ Acc@1\end{tabular}} \\ \hline
AlexNet          & 233.08 & 61.1  & 56.522 \\
Resnet-34        & 83.28  & 21.79 & 73.314 \\
Densenet121      & 30.99  & 7.97  & 74.434 \\
MobilenetV3      & 21.12  & 5.48  & 75.274 \\
EfficientNet\_B2 & 35.2   & 9.1   & 80.608 \\
\hline
\end{tabular}
\caption{Ensemble Members Details. We choose models based on their number of parameters and learning capacity. We use the Imagenet top-1 accuracy metric as our measure of learning capacity.}
\label{tab:arch-comp}
\end{table}

\subsection{Thief Dataset}
As our task is to extract Image Classification models, we use the complete ILSVRC-2012 challenge dataset (1.2M images) \cite{deng2009imagenet} as our attack dataset, as per previous work. This is similar to an actual attack scenario, as the attacker might scrape the web to collect images. The Imagenet is a comprehensive dataset in terms of the number of classes and image resolution. Appropriate model-specific transformations are applied to each image before inputting to the victim and thief models. 

\subsection{Training the ensemble}
For a valid comparison, we limit our maximum query budget to 30K, similar to previous works. 10\% of the budget is set aside as the validation set whose accuracy is monitored while training the ensemble. The number of query cycles is set to 10, allowing us 2.7K queries per cycle. That means every cycle, we select 2.7K samples out of the 1.2 million ImageNet images to query the victim model and then train our ensemble using the victim's output. The initial selection is done at random.

For all ensemble members, during the active learning stage, we use the SGD optimizer with a momentum of 0.9 for 200 epochs. The learning rate is set to 0.01 for Alexnet and 0.02 for others. The learning rate is decayed by a factor of 0.1 after every 30 epochs. Weight decay is set to 0 for all datasets. After ten cycles, each ensemble member has seen 27K samples chosen via the specified strategy. This leaves us with roughly 1.25 million samples for semi-supervised learning.

During the semi-supervised learning stage, we impose a selection limit of 100 samples for each class to avoid class imbalance. The learning rate is set to 0.002 to prevent catastrophic forgetting. Each epoch in this stage consists of a pass over the labelled as well as the unlabeled set. 

\subsection{Ensemble output and Comparison}
Majority voting is used as the ensemble's final decision-making strategy, i.e. if at least three models output the same label, it is considered the ensemble's output label.
We compare our method with ActiveThief\cite{pal2020activethief} and Black-Box Dissector (BBD) \cite{wang2022black}. We focus on the accuracy (Acc) and agreement (Agr) metrics as proposed by \cite{pal2020activethief} for each ensemble member and the ensemble as a whole. As test sets for the victim dataset are seldom available in a real-world setting, we pick the best model for each ensemble member based on its validation set accuracy in each cycle. Since none of the previous works utilizes our chosen thief model architectures for extraction, their accuracies are reported only for the scenario where the thief model architecture is the same as the victim model.

\section{Results and Analysis}

\begin{figure*}[]
    \centering
    
    \begin{subfigure}[b]{\textwidth}
        \centering
        \includegraphics[width=\textwidth]{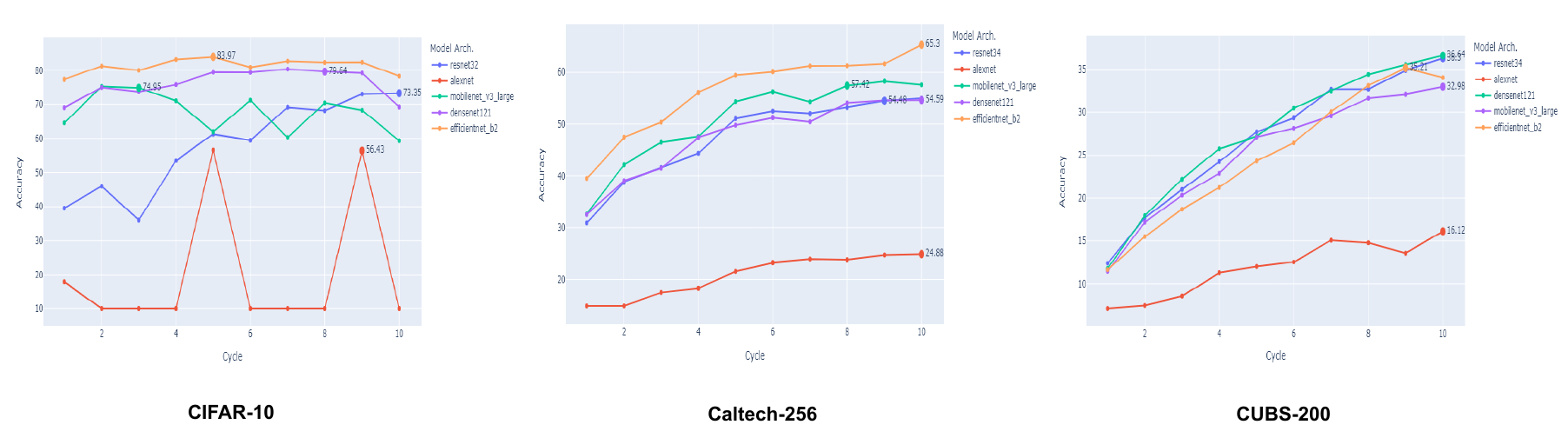}
        \caption{Test set accuracy curves for Consensus Entropy.}
        \label{fig:acc_consensus}
    \end{subfigure}
    \hfill
    \begin{subfigure}[b]{\textwidth}
        \centering
        \includegraphics[width=\textwidth]{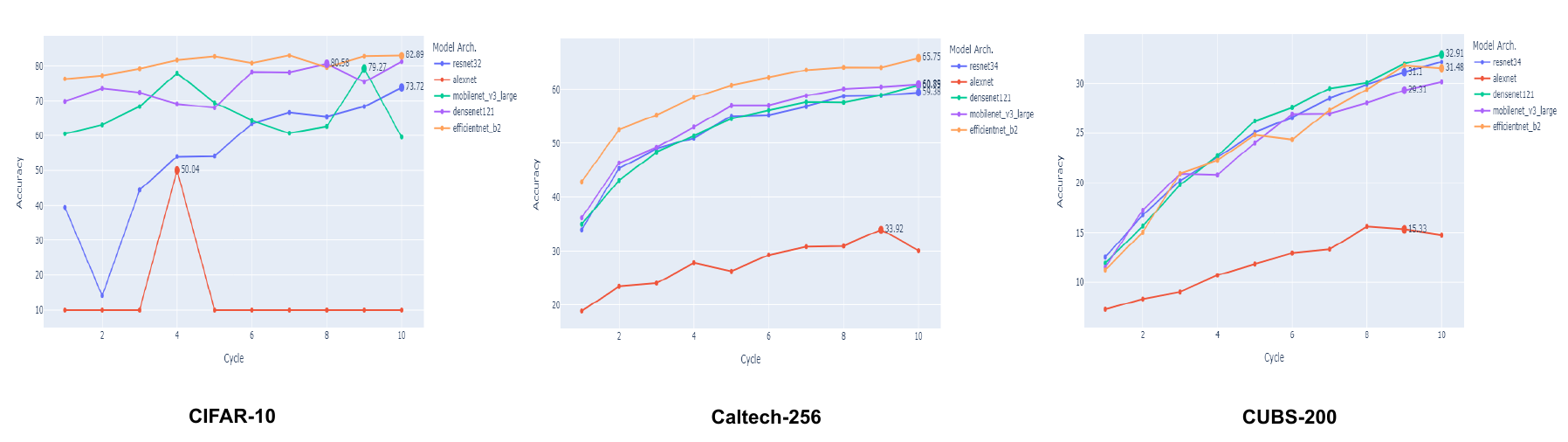}
        \caption{Test set accuracy curves for Label Disagreement.}
        \label{fig:acc_vote}
    \end{subfigure}
    \captionsetup[]{justification=centering}
    \caption{Accuracy of ensemble members in each cycle of the Active Learning process for a)Consensus Entropy and b) Label Disagreement on all selected datasets. AlexNet shows sudden spikes in accuracy for the CIFAR-10 dataset but is stable for other datasets. The best accuracy for each model is mentioned on the corresponding scatter line. We choose the best model based on the validation set accuracy as test sets aren't available in a real-world scenario.}
    \label{fig:accs}
\end{figure*}

\begin{table}[t!]
\centering
\resizebox{\columnwidth}{!}{%
\begin{tabular}{lcccccccc}
\hline
\multirow{2}{*}{Method} & \multicolumn{2}{c}{\textbf{CIFAR-10}} & \multicolumn{2}{c}{\textbf{CIFAR-100}}
& \multicolumn{2}{c}{\textbf{Caltech-256}} & \multicolumn{2}{c}{\textbf{CUBS-200}} \\ \cline{2-9} 
                    & Acc & Agr & Acc & Agr & Acc & Agr & Acc & Agr \\ \hline
Random Selection    & 75.64 & 76.88 & 43.0   & 42.78 & 59.01  & 61.17 & 33.78 & 36.9           \\
ActiveThief(Entropy)& 74.21 & 75.26 & 40.59  & 40.8 & 54.14  & 56.28 & 29.43 & 32.05            \\
ActiveThief(Kcenter)& 74.24 & 75.71 & 42.29   & 42.77    & 58.84  & 61.19 & 34.64 & 37.68       \\
BBD & \textbf{80.47}& \textbf{82.14}& 14.32  & 15.37 & 61.41  & 63.61 & 36.28 & 39.07           \\
+Kcenter & 79.27 & 80.84 & 12.2   & 13.38  & \textbf{63.75} & \textbf{66.34} & \textbf{44.43} & \textbf{48.46}   \\ \hline
AOT Consensus & 81.9        & 82.97 & 48.62  & 49.9  & 62.76  & 64.92 & 38.83 & 41.68         \\
+SSL          & 82.5        & 83.62 & \textbf{50.09}  & \textbf{51.76} & 60.81  & 63.03 & 39.29 & 42.19 \\
+KCenter       & 81.59  & 82.64 & 49.68  & 50.44 & \textbf{68.21}  & \textbf{69.65} & \textbf{45.85} & \textbf{50.17} \\
AOT Voting    & 82.93   & 83.82 & 47.86  & 47.95 & 64.75 & 66.1 & 33.32 & 36.46            \\
+SSL & \textbf{83.06} & \textbf{84.12} & 49.9   & 49.85 & 62.57       & 63.78         & 30.79 & 33.58           \\ \hline
\end{tabular}%
}
\captionsetup{justification=justified}
\caption{Experimentation Results. Accuracy (Acc) and Agreement (Agr) of the thief model are reported for each dataset. The highest results for our method and previous works are highlighted.}
\label{tab:main_results}
\end{table}

\subsection{Performance comparison}
Table \ref{tab:main_results} shows the results of our method in comparison with previous methods. We call our attack the Army of Thieves (AOT) attack. The results for the two sample selection strategies, consensus entropy, and label disagreement, are shown separately. As semi-supervised learning (SSL) was applied after budget exhaustion, the results are shown in separate rows. From the table, it is evident that our method outperforms previous work in terms of accuracy and agreement on CIFAR-10 and CIFAR-100. We achieved a higher accuracy on Caltech-256, and our agreement value is on par with previous work. We perform better than the base approach of BBD on CUBS-200 but fall short when they use the KCenter algorithm. We infer this is due to the inherent complexity of the CUBS-200 dataset, which is also evident from the low extraction accuracies of all previous approaches, the best being 44\%. The graphs showing each ensemble member's accuracy per active learning cycle are given in figures \ref{fig:acc_consensus} and \ref{fig:acc_vote}.

\subsection{Performance Analysis}
In this section, we study the performance of our attack in terms of accuracy and agreement metrics.
For CIFAR-10 and Caltech-256, we obtain the best performance when the selection strategy is disagreement-based instead of consensus-based. We observe that the semi-supervised learning approach helps the CIFAR-10 ensemble but fails for Caltech-256. We hypothesize that the class imbalance of the selected samples adversely affects the model. Due to the high number of classes (256), more samples get selected for the few highly confident classes in Caltech-256 as compared to CIFAR-10, which has only ten classes that are almost uniformly distributed. Due to the class imbalance, the model forgets information relevant to the less prevalent classes and leads to lower performance.

From the accuracy curves, it can be seen that accuracy for some models drops as more samples are selected. This contradicts the belief in active learning that accuracy should improve as more informative samples are selected with each cycle. We argue this happens primarily due to the nature of our ensemble, as a selected sample might not be equally informative for all models. This leads to minor drops in accuracy in the later stages of active learning. We also observe that Alexnet fails to learn optimally as the amount of available data is insufficient.
\begin{table}[]
\centering
\resizebox{\columnwidth}{!}{%
\begin{tabular}{llcc}
\hline
\textbf{Dataset}                                  & \textbf{Model Arch.}        & \textbf{Max Accuracy} & \textbf{Max Agreement} \\ \hline
\multicolumn{1}{l|}{\multirow{5}{*}{CIFAR-10}}    & \textbf{Alexnet}            & 63.93                 & 64.31                  \\
\multicolumn{1}{l|}{}                             & \textbf{DenseNet121}        & 80.55                 & 81.3                   \\
\multicolumn{1}{l|}{}                             & \textbf{EfficientNet\_B2}   & 87.61                 & 87.12                  \\
\multicolumn{1}{l|}{}                             & \textbf{MobileNetV3\_large} & 79.53                 & 80.62                  \\
\multicolumn{1}{l|}{}                             & \textbf{Resnet32}        & 74.21                 & 75.33                  \\ \hline
\multicolumn{1}{l|}{\multirow{5}{*}{CIFAR-100}}    & \textbf{Alexnet}           & 41.56                 & 43.06                  \\
\multicolumn{1}{l|}{}                             & \textbf{DenseNet121}        & 44.60                 & 45.19                  \\
\multicolumn{1}{l|}{}                             & \textbf{EfficientNet\_B2}   & 50.33                 & 48.68                  \\
\multicolumn{1}{l|}{}                             & \textbf{MobileNetV3\_large} & 43.62                 & 45.90                  \\
\multicolumn{1}{l|}{}                             & \textbf{Resnet34}        & 43.77                 & 44.36                  \\ \hline
\multicolumn{1}{l|}{\multirow{5}{*}{Caltech-256}} & \textbf{Alexnet}            & 41.87                & 43.45                  \\
\multicolumn{1}{l|}{}                             & \textbf{DenseNet121}        & 60.75                 & 62.43                  \\
\multicolumn{1}{l|}{}                             & \textbf{EfficientNet\_B2}   & 65.75                 & 63.92                  \\
\multicolumn{1}{l|}{}                             & \textbf{MobileNetV3\_large} & 60.9                  & 60.95                  \\
\multicolumn{1}{l|}{}                             & \textbf{Resnet34}        & 59.37                 & 61.93                  \\ \hline
\multicolumn{1}{l|}{\multirow{5}{*}{CUBS-200}}    & \textbf{Alexnet}            & 21.12                 & 22.36                  \\
\multicolumn{1}{l|}{}                             & \textbf{DenseNet121}        & 36.64                 & 39.52                  \\
\multicolumn{1}{l|}{}                             & \textbf{EfficientNet\_B2}   & 36.33                 & 37.31                  \\
\multicolumn{1}{l|}{}                             & \textbf{MobileNetV3\_large} & 33.7                  & 36.27                  \\
\multicolumn{1}{l|}{}                             & \textbf{Resnet34}        & 36.29                 & 39.16                  \\ \hline
\end{tabular}%
}
\caption{Individual model best accuracies}
\label{tab:best_accs}
\end{table}

For comparison with the ensemble, max accuracy and agreement metrics are reported for each individual model in Table \ref{tab:best_accs}. we observe that EfficientNet alone outperforms the ensemble for the CIFAR-10 and Caltech-256 datasets. Due to the differences in architectures of the victim and thief models, it becomes hard to verify whether the thief model has captured the exact decision boundaries as the victim or learnt entirely new ones due to the inherent capability of the architecture. We attempt to verify this by evaluating the adversarial sample transferability of each individual thief model. Adversarial samples created using MobileNet and EfficientNet do not transfer well to the victim Resnet-34 architecture. Both models generalize well in terms of accuracy and agreement but are not a valid ``copy" of the victim model as they fail to capture its decision boundaries.

\subsection{Adversarial Sample Transfer}
Adversarial Sample Transferability is measured by counting how many failures are shared across the victim and thief model. As showcased in Table \ref{tab:adv_transfer}, our method achieves superior adversarial sample transferability than other methods when the thief model architecture matches the victim model. Even for DenseNet, the transferability is better than in previous work. All the adversarial samples were generated using the Projected Gradient Descent (PGD)\cite{madry2019deep} with a maximum $L_{\infty}$-norm of 8/255 as per previous work in the field \cite{wang2022black}. For EfficientNet, MobileNet, and AlexNet, the transferability numbers are 54.81\%, 52.41\%, and 59.66\%, respectively. We argue that the transferability is low when the thief architecture does not match the victim, as the decision boundaries learned differ from the victim Resnet-34. The models provide high accuracy and agreement but cannot capture decision boundaries similar to the victim model.

\begin{table}[h]
\centering 
\resizebox{0.8\columnwidth}{!}{%
\begin{tabular}{lccccc}
\hline
\multicolumn{1}{c}{\multirow{2}{*}{\textbf{Method}}} & \multicolumn{5}{c}{\textbf{Substitute's Architecture}}    \\ \cline{2-6} 
\multicolumn{1}{c}{}                                 & Resnet-32 & DenseNet \\ \hline
ActiveThief(k-center)    & 57.44\% & 60.72\% \\
ActiveThief(Entropy)    & 63.56\% & 62.05\% \\
BBD & 76.63\% & 66.96\% \\
Ours  & \textbf{97.64}\% & \textbf{71.64}\% \\
\hline
\end{tabular}%
}
\caption{Adversarial Sample Transfer for the CIFAR-10 dataset using PGD attack. The numbers for previous works have been directly taken from the BBD\cite{wang2022black} paper.}
\label{tab:adv_transfer}
\end{table}

\begin{table}[htp]
\centering
\resizebox{\columnwidth}{!}{%
\begin{tabular}{llll}
\hline
\textbf{Model} &
  \begin{tabular}[c]{@{}l@{}}Victim\\ (Resnet-32)\end{tabular} &
  \begin{tabular}[c]{@{}l@{}}Thief\\ (Resnet-32)\end{tabular} &
  \begin{tabular}[c]{@{}l@{}}Thief\\ (DenseNet121)\end{tabular} \\ \hline
Clean             & 92.18          & 73.54 & 79.64 \\ \hline
PGD (Resnet-32)   & \textbf{0.6}   & 0 (97.64)      & -     \\
PGD (DenseNet121) & \textbf{24.43} & -     & 0 (71.64)     \\ \hline
BIA (Resnet152)   & 53.76          & 46.82 (61.60) & 48.04 (64.58) \\
BIA (DenseNet169) & 60.55          & 48.67 (55.22) & 51.55 (58.68)
\end{tabular}%
}
\caption{We report the top-1 accuracy after performing adversarial attacks on models trained on the CIFAR-10 dataset. The adversarial transferability is also reported in parenthesis. PGD attacks on our thief models are more effective as compared to the BIA attack.}
\label{tab:bia_attack}
\end{table}

We also compare our transferability with the cross-domain Beyond ImageNet Adversarial Attack (BIA) \cite{zhang2022imagenet} in Table \ref{tab:bia_attack}. We generate adversarial versions of each sample in the CIFAR-10 using pretrained generators provided by the authors. From the table, it is evident that the samples generated using the PGD attack on our thief models transfer better to the victim model and cause larger drops in accuracy. The best transferability is again observed when the thief and victim architectures are similar.

\subsection{Qualitative Analysis}
\begin{figure}
    \centering
    
    \begin{subfigure}[b]{0.4\textwidth}
        \centering
        \includegraphics[width=\textwidth]{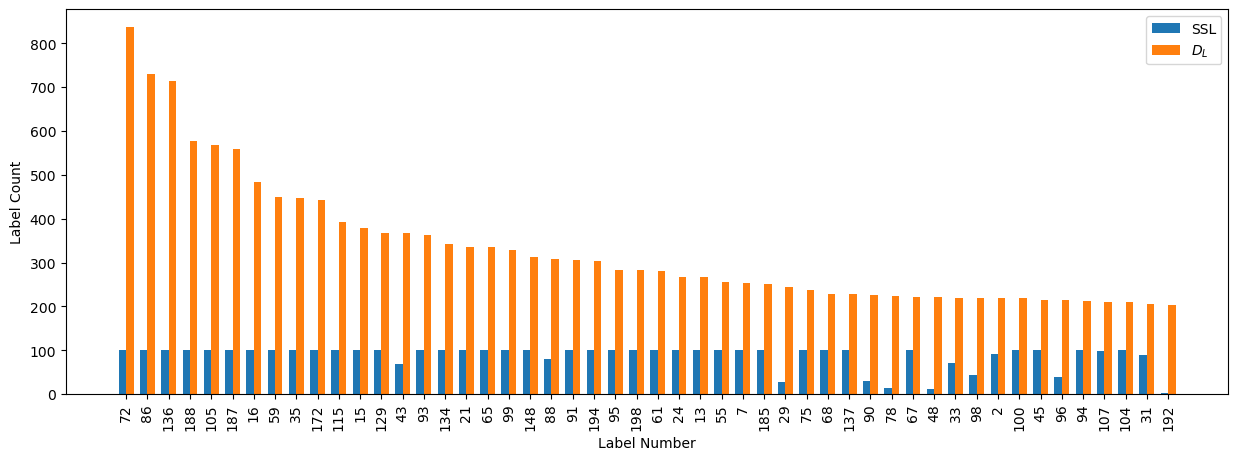}
        \caption{}
        \label{fig:subfig1}
    \end{subfigure}
    \hfill
    \begin{subfigure}[b]{0.4\textwidth}
        \centering
        \includegraphics[width=\textwidth]{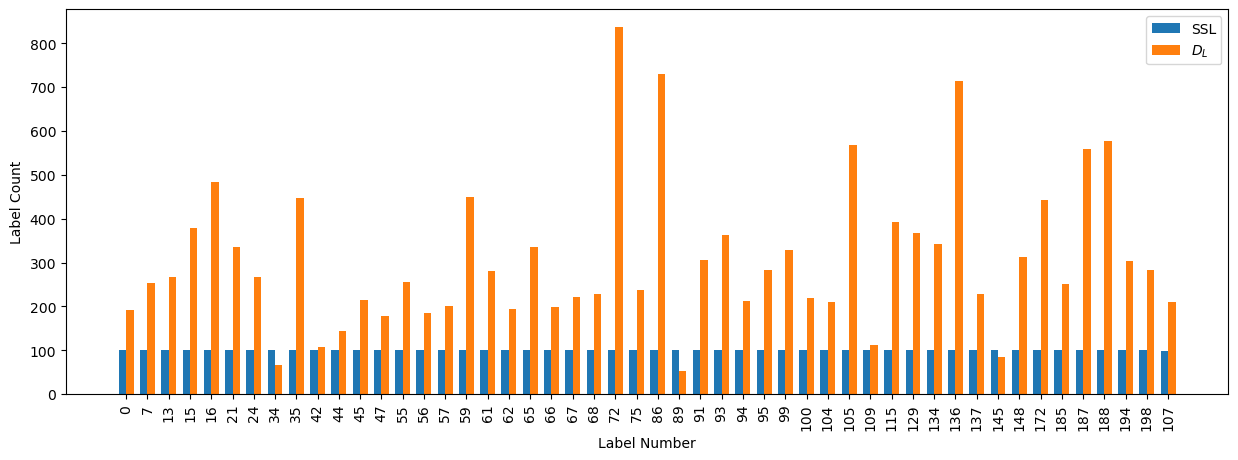}
        \caption{}
        \label{fig:subfig2}
    \end{subfigure}
    \captionsetup[]{justification=centering}
    \caption{a) Labels with highest frequencies in the Labeled set are selected as part of the Semi-supervised Learning set. b) Labels with high frequency in the Semi-supervised set almost always have high frequencies in the Labeled set. Both graphs are for the CUBS-200 dataset.}
    \label{fig:ssl_analysis}
\end{figure}

During semi-supervised learning, the selection of samples is typically based on the high confidence exhibited by the model towards certain classes. This approach establishes a direct relationship between the number of samples chosen per class through active learning and the model's confidence in those samples belonging to the respective class. We showcase this relationship for the CUBS-200 dataset in Figure \ref{fig:ssl_analysis}. Consequently, an overemphasis on these particular classes can occur, as they are repeatedly selected for semi-supervised learning, essentially reinforcing the existing knowledge of the model. Unfortunately, this leads to the loss of information pertaining to other classes, consequently resulting in a decline in the overall accuracy of the model. 

\section{Conclusion}
We explore the usage of an ensemble of thief models for the task of model extraction in the hard-label setting as a solution to the problems of noisy sample selection and the overfitting of thief models. Our method, called Army of Thieves, performs better than previous methods in terms of extracted accuracy and query efficiency. Even though we focus only on Image Classification in this work, our ensemble approach is generic enough to be applied to any domain like text, speech etc., given the presence of appropriate active and semi-supervised learning methods. The selection of ensemble members can be streamlined by an effective metric that quantifies the information-capturing ability of the architecture for the downstream task. We leave this task to future work and invite other researchers to work in this direction.

\section{Limitations and Discussions}
The foremost consideration when forming an ensemble of thief models is the selection of appropriate models. In our study, we opted for architectures with varying sizes and complexities. However, due to computational resource constraints, we could not include larger models such as Vision Transformer (ViT) and InceptionNet in our ensemble. Furthermore, the maximum ensemble size was limited to five members. It is worth noting that in the context of replicating the victim model, the size or efficiency of the models does not significantly impact the outcome. While larger models may possess the ability to capture more information and exhibit improved generalization, they do not serve a purpose from an adversarial perspective. Ultimately, the primary objective of replicating the victim model is not effectively achieved by utilising larger models. 

Prior studies in model extraction have incorporated the k-centre or core-set method alongside their proposed techniques to enhance active learning on extensive datasets. Our algorithm surpasses previous methodologies that employ the k-centre method on CIFAR-10, CIFAR-100 and Caltech-256 datasets. Upon incorporation of k-centre algorithm into our pipeline, our approach outperforms all previous SOTA approaches on the CUBS-200 dataset as well. We postulate that the performance increase of ~6.5\% stems from the unique characteristics of the CUBS-200 dataset, where distinctions between classes are subtle, necessitating fine-grained classification. In the end, we would still argue that employing an ensemble approach generally yields superior results for model extraction.


\section*{Acknowledgements}
The authors acknowledge the support of Infosys Centre for AI (CAI) at IIIT-Delhi and iHub-Anubhuti-IIITD Foundation set up under the NM-ICPS scheme of the DST.

{
\small
\bibliographystyle{ieee_fullname}
\bibliography{citations}
}

\end{document}